\def\year{2021}\relax
\documentclass[letterpaper]{article} 
\usepackage{aaai18}  
\usepackage{times}  
\usepackage{amsmath}
\usepackage{helvet}  
\usepackage{courier}  
\usepackage{url}  
\usepackage{graphicx}  

\begin{document}
%
\title{Do Fine-tuned Commonsense Language Models Really Generalize?}
\author{Mayank Kejriwal, Ke Shen\\
Information Sciences Institute\\
USC Viterbi School of Engineering\\
4676 Admiralty Way 1001\\
Marina Del Rey, California 90292\\
}
\maketitle
\begin{abstract}
Recently, transformer-based methods such as RoBERTa and GPT-3 have led to significant experimental advances in natural language processing tasks such as question answering and commonsense reasoning. The latter is typically evaluated through multiple benchmarks framed as multiple-choice instances of the former. According to influential leaderboards hosted by the Allen Institute (evaluating state-of-the-art performance on commonsense reasoning benchmarks), models based on such transformer methods are approaching human-like performance and have average accuracy well over 80\% on many benchmarks. Since these are \emph{commonsense} benchmarks, a model that \emph{generalizes} on commonsense reasoning should not experience much performance loss across multiple commonsense benchmarks. In this paper, we study the generalization issue in detail by designing and conducting a rigorous scientific study. Using five common benchmarks, multiple controls and statistical analysis, we find clear evidence that fine-tuned commonsense language models still do not generalize well, even with moderate changes to the experimental setup, and may, in fact, be susceptible to dataset bias. We also perform selective studies, including qualitative and consistency analyses, to gain deeper insight into the problem.
\end{abstract}

\section{Introduction}
Commonsense reasoning has become a resurgent area of research in both the NLP and broader AI communities\footnote{Two other example domains include computer vision \cite{zellers2019recognition} and social networks \cite{dinakar2012common}.} \cite{res2}, \cite{res3}, \cite{storks2019}, despite having been introduced as an early AI challenge more than 50 years ago (in the context of machine translation) \cite{MT1960}. Traditionally, it was believed that the problem could only be solved through a combination of techniques, including Web mining, logical reasoning, handcrafted knowledge bases and crowdsourcing \cite{marcus2015}, \cite{liu2004conceptnet}, \cite{moore1982role}. More recently, the advent of powerful `transformer' neural networks, especially in NLP \cite{BERT}, \cite{liu2019roberta}, suggests that the time is right to build commonsense reasoners that generalize to a wide variety of situations, including those involving social and physical reasoning \cite{socialiqa}, \cite{piqa}. There are several related reasons why commonsense reasoning is such an important topic in AI. Commonsense reasoning is an innately human ability that machines have (thus far) not proven adept at `conquering' unlike other task-specific domains such as face recognition \cite{liu2017sphereface}. Perhaps for that reason, it has always presented an enticing challenge to many AI researchers throughout the decades \cite{CS1}, \cite{CS2}, \cite{CS3}, \cite{CS4}. There is also the widely held belief that, for a `general AI' to truly emerge, commonsense reasoning is one problem (among others) that will need to be solved in a sufficiently robust manner \cite{generalAI}. A more functional reason for increased interest in commonsense reasoning is the rise of chatbots and other such `conversational AI' services (e.g., Siri and Alexa) that represent an important area of innovation in industry \cite{convo1}, \cite{convo2}, \cite{convo3}, \cite{convo4}. Recently, the US Department of Defense also launched a machine common sense (MCS) program in which a diverse set of researchers and organizations, including the Allen Institute of Artificial Intelligence, is involved \cite{MCS}.    

Despite the success of these models, there is some evidence (not necessarily all quantitative) to suggest that the models are still superficial i.e. do not have the same commonsense abilities as humans, despite what the performance numbers suggest. \cite{marcus2015} suggested in a seminal review article that, for truly human-level, performance `knowledge of the commonsense world--
time, space, physical interactions, people,
and so on—will be necessary.' While we do not deny the theoretical possibility that a language representation model such as BERT, RoBERTa or GPT-3 may have learned these different aspects of the real world purely by `reading' large corpora of natural language \cite{BERT}, \cite{liu2019roberta}, \cite{GPT3}, we do claim that such possibilities can (and must) be tested through rigorous evaluation. Unfortunately, as we cover in the \emph{Related Work}, there has been little to no work by way of conducting such a systematic and focused analysis (with the central goal of evaluating generalization of a system on commonsense reasoning), using a publicly available and replicable system, though there is plenty of precedent for this type of study, as discussed in the \emph{Related Work}. 

In this paper, we attempt to address this gap by carefully designing and conducting an empirical study with the specific intent of answering the question of whether fine-tuned commonsense language models generalize in robust ways. Our goal is not to attack either a model or a particular benchmark (or a set of benchmarks) but to present clear (and cautionary) evidence that the current set of evaluations (and evaluation practices) and reported results need to be considered with more skepticism by the community. Considering the pace at which research on commonsense reasoning continues, we posit that this is a timely study and could serve as a methodology for future such studies assessing the generalization of commonsense AI.


\section{Related Work}
As noted in the \emph{Introduction}, commonsense reasoning has recently experienced a resurgence in the AI research community. Central references that attest to this resurgence include \cite{res1}, \cite{res2}, \cite{res3}, \cite{res4}, \cite{res5}, \cite{zellers2019recognition}. We also noted that commonsense reasoning has also been an ambitious agenda in the past. It is not feasible to cite all relevant work herein; instead, we refer the reader both to the review article by \cite{marcus2015}, as well as more recent surveys on commonsense reasoning tasks and benchmarks \cite{storks2019}.

Much progress has been made on specific \emph{kinds} of commonsense reasoning, especially in reasoning about time and internal relations \cite{taxonomy1}, \cite{Pinto}, reasoning about actions and change \cite{Srinivas}, and the sign calculus \cite{marcus2015}. Semantics have played an important role in some of these successes \cite{semantics1} including `semantic networks' and other structured sources, important ones of which include ConceptNet  \cite{conceptnet3}. WordNet \cite{wordnet} and Cyc \cite{Cyc}. These resources have been frequently applied in multiple reasoning systems \cite{botschen}, \cite{angeli}, \cite{Lin2017}. In contrast with WordNet and ConceptNet, Cyc focuses on designing a universal schema (a higher-order logic) to represent commonsense assertions, which also supports reasoning systems \cite{Panton06commonsense}, \cite{Ramachandran05}  conduct richer logical inference. 

In the last decade, in particular, as more accurate but also more `non-interpretable' (or explainable) models like neural networks have become more prevalent, a relevant line of research has developed in `adversarially attacking' these models to understand their weaknesses in a variety of domains \cite{adv1}, \cite{adv2}, \cite{adv3}. Other problems, that require more precise inputs and prompts, include bias in the data and also in the model \cite{bias1}, \cite{bias2}. This line of work is valuable precedent for our own work, and there has been some early work already on conducting such robustness tests on transformer-based language representation models \cite{tadv1}, \cite{tadv2}, \cite{tadv3}. However, this paper significantly departs in at least one respect from these lines of work--namely, we do not adversarially or selectively modify the input or the model in any way. Our results show, in fact, that sophisticated adversarial modifications are not necessary for concluding that generalization is a concern for transformer-based QA models.

Theoretical work on commonsense reasoning along the lines of cognitive science and computational commonsense paradigms should also be noted \cite{hobbs1}, \cite{hobbs2}, \cite{hobbs3}. We note this line of work because it could potentially be used for designing better evaluations, as well as for diagnosing why some transformer-based models are not generalizing better, despite (individually) good performance across the board on many benchmark datasets.

\section{Background and Preliminaries}

\subsection{Commonsense Question Answering (QA) Benchmarks}
\begin{figure}
  \centering
  \includegraphics[width=3.5in]{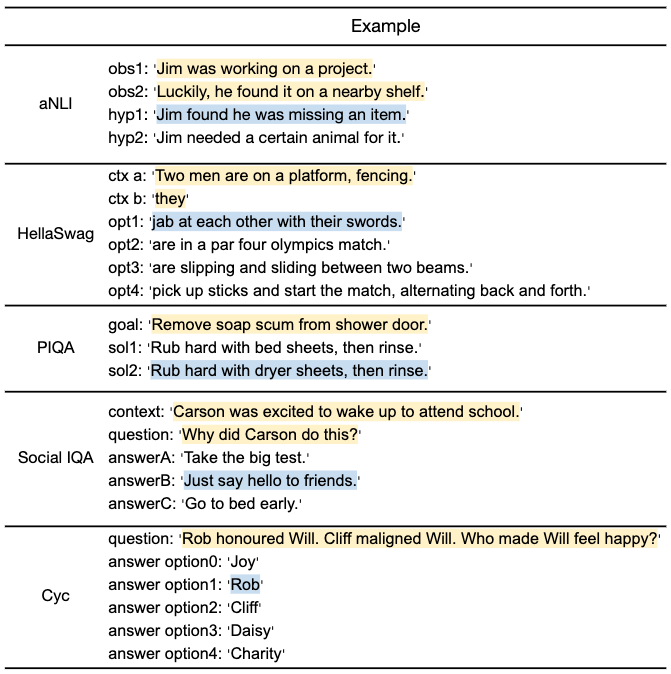}
  \caption{Question-answer instances from five commonsense benchmark datasets used for the evaluations in this paper. The question-like `prompt' is highlighted in yellow, and the correct answer in blue.} 
  \label{figure: conNet_frag}
\end{figure}
As noted in both the introduction and the related work, commonsense reasoning has emerged as an important and challenging research agenda in the last several years. The usual way to evaluate systems (with the state-of-the-art systems being based, in some significant way, on the transformer-based models described in the next section) purporting to be capable of commonsense reasoning in the natural language setting is to use \emph{question answering} benchmarks with multiple choices per `question' from which exactly one correct answer must be selected by the system. The NLP (and broader AI) community has developed numerous such benchmarks, especially in the last 3-5 years, using a range of methodologies both for acquiring the questions and for devising the answers. We describe the five benchmarks used in the research study in this paper below, with references for further reading. Examples are provided in Figure \ref{figure: conNet_frag}.

\begin{enumerate}
    \item {\bf aNLI (Abductive Natural Language Inference):} Abductive Natural Language Inference (aNLI)\footnote{\url{https://storage.googleapis.com/ai2-mosaic/public/alphanli/alphanli-train-dev.zip}} \cite{anli} is a new commonsense benchmark dataset designed to test an AI system’s capability to apply abductive reasoning and common sense to form possible explanations for a given set of observations. Formulated as a binary-classification task, the goal is to pick the most plausible explanatory hypothesis given two observations from narrative contexts. 
    \item {\bf HellaSwag:} HellaSWAG\footnote{\url{https://storage.googleapis.com/ai2-mosaic/public/hellaswag/hellaswag-train-dev.zip}} \cite{hellaswag} is a dataset for studying grounded commonsense inference. It consists of 70,000 multiple choice questions about `grounded situations': each question comes from one of two domains--Activity Net or wikiHow--with four answer choices about what might happen next in the scene. The correct answer is the (real) sentence for the next event; the three incorrect answers are adversarially generated and human-verified, ensuring a non-trivial probability of `fooling' machines but not (most) humans.
    \item {\bf PIQA:} Physical Interaction QA (PIQA)\footnote{\url{https://storage.googleapis.com/ai2-mosaic/public/physicaliqa/physicaliqa-train-dev.zip}} \cite{piqa} is a novel commonsense QA benchmark for naive physics reasoning, primarily concerned with testing machines on how we interact with everyday objects in common situations. It tests, for example, what actions a physical object `affords' (e.g., it is possible to use a cup as a doorstop), and also what physical interactions a group of objects afford (e.g., it is possible to place an computer on top of a table, but not the other way around). The dataset requires reasoning about both the prototypical use of objects (e.g., glasses are used for drinking) but also non-prototypical (but practically plausible) uses of objects. 
    \item {\bf Social IQA:} Social Interaction QA\footnote{\url{https://storage.googleapis.com/ai2-mosaic/public/socialiqa/socialiqa-train-dev.zip}} \cite{socialiqa} is a QA benchmark for testing social common sense. In contrast with prior benchmarks focusing primarily on physical or taxonomic knowledge, Social IQA is mainly concerned with testing a machine's reasoning capabilities about people’s actions and their social implications. Actions in Social IQA span many social situations, and answer-candidates contain both human-curated answers and (`adversarially-filtered') machine-generated candidates. 
    \item{\bf CycIC:} Cyc Intelligence Challenge dataset (Cyc)\footnote{\url{https://storage.googleapis.com/ai2-mosaic/public/cycic/CycIC-train-dev.zip}} is a set of multiple choice and true/false questions requiring general common sense knowledge and reasoning in a very broad variety of areas, including simple reasoning about time, place, everyday objects, events, and situations. Some questions may require some logic to get the correct answer. Here, we only use the multiple-choice questions (and not true/false questions) for experiments.
\end{enumerate}

One important aspect to note about these benchmarks is that, while all offer multiple-choice answer formats, the `prompt' is not always a question. For example, in the case of the aNLI benchmark, the `prompt' is a set of two observations, and the `choices' are two hypotheses (of which the one that best fits these observations should be selected). For this reason, we refer to each question and corresponding answer choices as an \emph{instance}. The instance formats of the benchmarks described above are stated in Table \ref{tab:data_format}.    

\begin{table}
\caption{`Instance formats' of commonsense QA benchmarks. The -\textgreater{} is used to separate what is given (e.g., obs1, obs2 for aNLI) and the answer choices (hypo1, hypo2). Here, `obs', `hypo', `opt', `ans', `sol' and `ctx' stand for \emph{observation, hypothesis, option, answer, solution} and \emph{context} respectively.}
\begin{center}
\begin{tabular}{|l|l|}
\hline
             & Format\\ \hline
aNLI         & obs1, obs2-\textgreater{}hypo1, hypo2\\ \hline
HellaSwag    & \begin{tabular}[c]{@{}l@{}}ctx\_a, ctx\_b-\textgreater{}opt1, opt2, \\ opt3, opt4\end{tabular} \\ \hline
PIQA & goal-\textgreater{}sol1, sol2\\ \hline
Social IQA   & context, question-\textgreater ans1, ans2, ans3 \\ \hline
Cyc        & question-\textgreater{}ans0, ans1, ans2, ans3, ans4\\ \hline
\end{tabular} 
\end{center}\label{tab:data_format}
\end{table}


\subsection{Transformer-based Models and RoBERTa}

As covered in the \emph{Related Work}, transformer-based models have rapidly emerged as state-of-the-art in the natural language processing community, both for specific tasks like question answering, but also for deriving `contextual embeddings' (for more details, we refer the reader to the citations in that section). RoBERTa has been a particularly successful model, and is (in essence), a highly optimized and better-trained version of BERT. Unlike the most recent model (GPT-3), a pre-trained version of RoBERTa is fully available for researchers to use and can be `fine-tuned' for specific tasks \cite{liu2019roberta}. Unsurprisingly, many of the systems occupying the top-5 leaderboard positions\footnote{\url{https://leaderboard.allenai.org/}.} for the commonsense reasoning benchmarks described earlier are based on RoBERTa in some significant manner. The experiments in this paper, described next, use a publicly available \emph{RoBERTa Ensemble} model\footnote{\url{https://github.com/isi-nlp/ai2/tree/base}} that was not developed by any the authors, either in principle or practice, can be downloaded and replicated very precisely, and on average, achieves over 80\% on the five benchmarks when fine-tuned on the benchmark without any change to the model itself.   

\section{Experiments}

We design and conduct a rigorous series of experiments (with full statistical analyses) to study the question noted in the title of this paper itself. While the data and system have already been described in the previous section, we use the next section to provide some relevant technical and methodological details, followed by the results.  

\subsection{Data and Methodology}

We use the five benchmarks described earlier for our evaluation datasets. Each of these benchmarks is publicly available, and even has a leaderboard dedicated to it. Many researchers have used these benchmarks for evaluating commonsense reasoning models \cite{storks2019}. Note that the `test' partition of these benchmarks is not available publicly; hence, for research purposes, the development or `dev.' set is used as the test set. To ensure replication, we do the same. Our goal here is not to develop a superior algorithm that may do better on the unknown test set, but to explore the capabilities of a popular language model-based solution to this problem. Details on the benchmarks' training and development set partitions, as well as current state-of-the-art (SOTA) performance by a highly optimized RoBERTa system on the leaderboard (designed and fine-tuned on just that specific task or benchmark) are shown\footnote{These numbers were acquired from the Allen Institute's leaderboards in late August, and may have shifted slightly.} in Table \ref{tab:performance_size}. As described in  the previous section, we used the \emph{RoBERTa Ensemble} model for our experiments, which achieves over 80\% performance (on average) over the five benchmarks and is not substantially different from  the SOTA model. While some formatting was necessary in order to ensure that the \emph{RoBERTa Ensemble} system, when trained on one dataset (say Cyc), could be applied to instances from another dataset (say Social IQA), we did not modify any of the information content within the instances (either in the questions/prompts or in the answers). 

\begin{table}[]
\caption{State-of-the-art (SOTA) performance and number (num.) of instances in training (train.) and development (dev.) partitions of five commonsense QA benchmarks.}
\begin{center}
\begin{tabular}{|p{0.7in}|p{0.75in}|p{0.7in}|p{0.5in}|}
\hline
            Benchmark & Num. train. set instances & Num. dev. set instances & SOTA accuracy \\ \hline
aNLI         & 169,654        & 1,532        & 0.873                    \\ \hline
HellaSwag    & 39,905         & 10,042       & 0.939                    \\ \hline
PIQa & 16,113         & 1,838        & 0.901                    \\ \hline
Social IQA   & 33,410         & 1,954        & 0.832                    \\ \hline
Cyc       & 6,570          & 947          & 0.913                    \\ \hline
\end{tabular}
\end{center}\label{tab:performance_size}
\end{table}
Furthermore, since one of our goals is to test the generalization ability of precisely such models (i.e. a `commonsense QA' model that has been trained on one kind of commonsense data, but evaluated on another), we define a \emph{performance loss metric (PL)} as: 
\begin{equation}
    PL = \frac{Acc_{indomain} - Acc_{outdomain}}{Acc_{indomain}}.
\end{equation}
 Here, $Acc_{indomain}$ is the `in-domain' prediction accuracy achieved on the benchmark when we  train the model on that benchmark's training set partition and evaluate on the development set from the \emph{same} benchmark; $Acc_{outdomain}$ is the `out-domain' prediction accuracy achieved when one benchmark is used for training and another benchmark's \emph{dev.} set partition is used for evaluation. Since there are four training options (the other four benchmarks) once the dev. benchmark has been fixed, it is possible to compute four separate $Acc_{outdomain}$ metrics for any given benchmark. The PL has an intuitive interpretation: how much of the `in-domain' performance (in percentage terms) does the model `lose' when the evaluation  benchmark is changed? Given the descriptions of the five benchmarks used in this paper in the previous section, we would \emph{expect} that the PL would be greatest for the benchmarks that are highly narrow in their domains (e.g., if we train on Social IQA and test on PIQA) as opposed to cases when the training benchmark is broad (such as when the training benchmark is aNLI, HellaSwag and Cyc). In the next section, we assess the validity of this hypothesis. Note that a high PL implies that the model is generalizing less. A negative PL is theoretically possible (and potentially unbounded from below, as $Acc_{indomain}$ tends to 0), but not observed. The PL can never be greater than 100\% (when $Acc_{outdomain}=0$).
 
Statistical significance is an important part of such studies. To ensure that results are valid, we conduct two kinds of significance testing for each `out-domain' experiment. Both tests use the `base' or in-domain setting (\emph{train.} and \emph{dev.} partitions come from the same benchmark during testing) as the `reference' distribution against which we test for equality of means. Specifically, we conduct the \emph{paired} Student's t-test between the in-domain accuracy data and the out-domain accuracy data when the \emph{dev.} set of the reference and the test distribution coincide (but the \emph{train.} does not). For example, for the experiment where we train on  aNLI, but test on HellaSwag, we conduct the paired Student's t-test between the out-domain accuracy data on HellaSwag and the in-domain accuracy data on HellaSwag (since the same \emph{dev.} set was used for collecting both sets of data). Since the results are `aligned',\footnote{The reason being that the `reference' (the model trained on HellaSwag) and the out-domain system (the model trained on aNLI) are both tested on the same questions, namely the dev. partition of HellaSwag.} the paired Student's t-test is applicable. 

The second test (the unpaired Student's t-test) is the converse of the above--namely, when the training sets coincide but the \emph{dev.} sets are different. Since the \emph{dev.} sets are different, we can only test for equality of means in the unpaired setting, since data points are not aligned across systems. Taking again the above example, we would conduct the unpaired Student's t-test between the out-domain result (trained on aNLI, but tested on HellaSwag) and the in-domain result achieved on aNLI (trained on aNLI, tested on aNLI). Note that both tests are complementary and necessary in this kind of experiment, since both the training and test settings can change. The first test evaluates significance of results holding the \emph{dev.} set fixed, while the second test keeps the training set fixed.   

\subsection{Results}
\begin{table}[h]
\caption{Accuracy (fraction of questions of which the answers are correctly predicted) of the \emph{Roberta Ensemble} model in different evaluation settings. The row represent the benchmark of which the training set partition is used to train, while the column name represents the benchmark of which the development (dev.) set partition is used to test. }
\begin{center}
\begin{tabular}{|l|c|p{0.3in}|c|p{0.3in}|c|}
\hline
\multicolumn{1}{|c|}{} & aNLI           & Hella Swag     & PIQA           & Social IQA     & Cyc            \\ \hline
aNLI                   & 0.819          & 0.611         & 0.702          & 0.531          & 0.442          \\ \hline
HellaSwag              & 0.681          & 0.835         & 0.699          & 0.515          & 0.351 \\ \hline
PIQA                   & 0.68           & 0.564         & 0.756          & 0.51           & 0.371          \\ \hline
Social IQA             & 0.688          & 0.604         & 0.687          & 0.769          & 0.508          \\ \hline
Cyc                    & 0.628 & 0.49 & 0.628 & 0.493 & 0.811          \\ \hline
\end{tabular}
\end{center}
\end{table}

\begin{table}[h]
\caption{The performance loss (PL; Equation 1) of the \emph{Roberta Ensemble} model when evaluated on a different benchmark (`out-domain') than it was trained on (`in-domain'). The row represent the benchmark of which the training set partition is used to train, while the column name represents the benchmark of which the development (dev.) set partition is used to test.}
\begin{center}
\begin{tabular}{|l|c|p{0.3in}|c|p{0.3in}|c|}
\hline
\multicolumn{1}{|c|}{} & aNLI           & Hella Swag     & PIQA           & Social IQA     & Cyc            \\ \hline
aNLI                   & 0                         & 0.268                 & 0.071            & 0.309               & 0.455        \\ \hline
HellaSwag              & 0.168                     & 0                              & 0.075                     & 0.330                           & 0.567           \\ \hline
PIQa                   & 0.169                     & 0.325                          & 0               & 0.336                           & 0.543                    \\ \hline
Social IQA             & 0.159           & 0.276                          & 0.09                      & 0                               & 0.374           \\ \hline
Cyc                    & 0.233            & 0.413                 & 0.169            & 0.358                  & 0                        \\ \hline
\end{tabular}
\end{center}
\end{table}

The absolute accuracy results of the \emph{RoBERTa Ensemble} model when trained on one benchmark and tested on  another are tabulated in Table 3. Overall, we see very clear evidence that, regardless of the \emph{train.} dataset used, out-domain performance inevitably declines, sometimes by significant margins. It is telling that these declines occur, regardless of whether we train on a `broader' commonsense benchmark (like HellaSwag) or whether we test on a broad or narrow benchmark (e.g., PIQA).  For better analysis of relative differences, we tabulate the performance loss (PL) metric in  Table 4. The diagonals are all 0, since the performance loss is 0 when the training and testing benchmark are the same (per Equation 1). The numbers correspond closely to those in Table 3, but generally tend in the opposite direction (i.e. PL is lower when the absolute accuracy is higher for a test benchmark, all else being the same.). 

Recall that, in the previous section, we stated a hypothesis that we expect test benchmarks that are too `narrow' (such as PIQA or Social IQA) to exhibit more PL than benchmarks which are broader, except (possibly) when the training set is also broad. Table 4 shows that the data on this question is surprisingly mixed. In particular, PL on PIQA is always low when it is used as a test set, despite the fact that it covers such a narrow domain. In contrast, the PL on Social IQA is high (usually, the second highest after Cyc). Similarly, with respect to testing on the `broader' benchmarks, PL is low on aNLI but higher on HellaSwag. When training on aNLI or HellaSwag, and comparing to training on either PIQA or Social IQA, we find that the difference is not considerable e.g., the system trained on HellaSwag achieves PL of 16.8\%, 33\% and 56.7\% respectively on aNLI, Social IQA and Cyc, and the system trained on PIQA achieves PL of 16.9\%, 33.6\% and 54.4\% respectively on the same three test sets. Therefore, it is simply not true that performance loss is observed simply because the `domains are different' (though by definition, they are all commonsense benchmarks), which is sometimes the cause in similarly designed (and more traditional) transfer learning and weak supervision experiments. 

Interestingly, based both on the data in Tables 3 and 4, we find clear evidence that Cyc is the most `different' benchmark, since the PL is markedly higher with Cyc used as the training (and also  the testing) dataset. Namely, the PLs observed in the Cyc `column' are the highest among the values in the corresponding training dataset's `row' e.g., when PIQA is used as the training dataset, the PL of 54.3\% observed for Cyc is the highest in that row.

{\bf Significance Analysis.} The paired Student's t-test methodology (the `first' significance test mentioned in the previous section) was first applied to all values in Table 3, and compared against the `diagonal' value. For example, the results obtained when training (respectively) on HellaSwag, PIQA, Social IQA and Cyc,  and testing on aNLI, are tested individually using the test statistic from a paired Student's t-test analysis against the the in-domain aNLI setting (the diagonal accuracy value of 81.9\% in Table 3). We find that the null hypothesis can be rejected at the 99\% level for all such paired tests, for all test benchmarks. The differences in accuracy are therefore significant, as are the PLs in Table 4, which is just a scaled, affine transformation of Table 3. 

We also conducted the unpaired Student's t-test (the `second' significance test mentioned in the previous section). For example, we compared the \emph{dev.} set results on HellaSwag, PIQA, Social IQA and Cyc (individually), when trained on aNLI, against the (\emph{dev.} set) results obtained on aNLI for the same model. Therefore, the training set (and hence, the model) is identical in all settings, but the \emph{dev.} set is different. The results for the unpaired Student's t-test showed that the majority of results are significant at the 99\% level; however, for the models trained on Social IQA and Cyc, we found that the (respective) results obtained when aNLI's \emph{dev.} set is used, are significant only at the 95\% level against the `reference' results obtained when the \emph{dev.} set of Social IQA and Cyc is used, respectively. A lone insignificant result (even at the 90\% level) is when we train on HellaSwag and test on aNLI, and compare it to the results of the same model tested on HellaSwag.

\begin{table}[h]
\caption{Consistency analysis of out-domain models' (ODMs) predictions on the questions on which the in-domain model got correct (top) and incorrect (bottom) predictions. The column name indicates the test benchmark. The sum of each column (in the respective table) is the number of questions on which the corresponding in-domain model gets correct/incorrect answers.}
\begin{tabular}{|p{1.1in}|p{0.3in}|p{0.3in}|p{0.25in}|p{0.3in}|p{0.2in}|}
\hline
\multicolumn{1}{|c|}{} & aNLI           & Hella Swag     & PIQA           & Social IQA     & Cyc            \\ \hline
All ODMs give the right answer            & 512  & 2,421      & 707  & 297        & 100 \\ \hline
Some ODMs give the right answer            & 683  & 5,033      & 630  & 1,043       & 503 \\ \hline
All ODMs give same wrong answer & 59   & 340       & 52   & 66         & 48  \\ \hline
Some ODMs disagree on wrong answers & 0    & 594       & 0    & 97         & 118 \\ \hline
\end{tabular}
\newline
\vspace*{0.2 cm}
\newline
\begin{tabular}{|p{1.1in}|p{0.3in}|p{0.3in}|p{0.25in}|p{0.3in}|p{0.2in}|}
\hline
\multicolumn{1}{|c|}{} & aNLI           & Hella Swag     & PIQA           & Social IQA     & Cyc            \\ \hline
All ODMs give the right answer            & 31   & 101       & 55   & 33         & 7 \\ \hline
Some ODMs give the right answer            &  170  & 957       & 266  & 296        & 89 \\ \hline
All ODMs give same wrong answer & 77   & 195       & 128  & 55         & 20  \\ \hline
Some ODMs disagree on wrong answers & 0    & 401       & 0    & 67         & 63  \\ \hline
\end{tabular}
\end{table}
{\bf Discussion on Observed Differences.} The previous results clearly illustrate that the choice of the training benchmark matters, often in surprising (but statistically significant) ways. One hypothetical reason why this behavior is observed is that PIQA may just be an `easier' dataset, and Cyc may just be a `harder' dataset (hence leading to lower and higher PLs respectively). However, if this were the case, then the `diagonal' values in Table 3 would reflect it. The observed values in Table 3 tell a different story; all results are clustered in  the range of 75-83\%, and the in-domain result for PIQA is similar to that of Social IQA, suggesting that the two are of reasonably equal difficulty. Yet, one proves to be more `generalizable' than another in out-domain settings. Another hypothetical reason could be the `number' of answer choices available. The hypothesis is that, once there is a mismatch between the training and testing setting, the performance becomes more `random'. While this hypothesis may explain why Cyc has the highest PL (it has 5 answer choices, generally, for every question; see Table 1), it does not explain why Social IQA (which has three answer choices per question) has higher PL than HellaSwag (which has four). Furthermore, the large differences in accuracy observed in the out-domain settings in  Table 3 cannot be explained only by differences in the number of answer choices available in  the different benchmarks. Finally, if the model had become more `random', expected accuracy on  Cyc and aNLI would be around 20\% and 50\% respectively (assuming relatively equal distribution of answer choices); however, the accuracies are significantly higher, according to Table 3. The model is clearly learning `something'. Furthermore, given the relatively large sizes and broad domains covered by some of  these datasets (see Table 2; even Cyc, the smallest `training' dataset has more than 6,000 questions in its training partition), it is unlikely that the fault lies purely with the data. The most plausible explanation that remains is that the fine-tuned RoBERTa model is subject to some degree of dataset bias, and that leaderboard performance numbers on individual benchmarks should not necessarily be assumed to be a reflection of advances in human-like `commonsense reasoning' without significantly more qualification.  
\begin{figure*}[t]
\centering
\includegraphics[height=3.4in, width=7.0in]{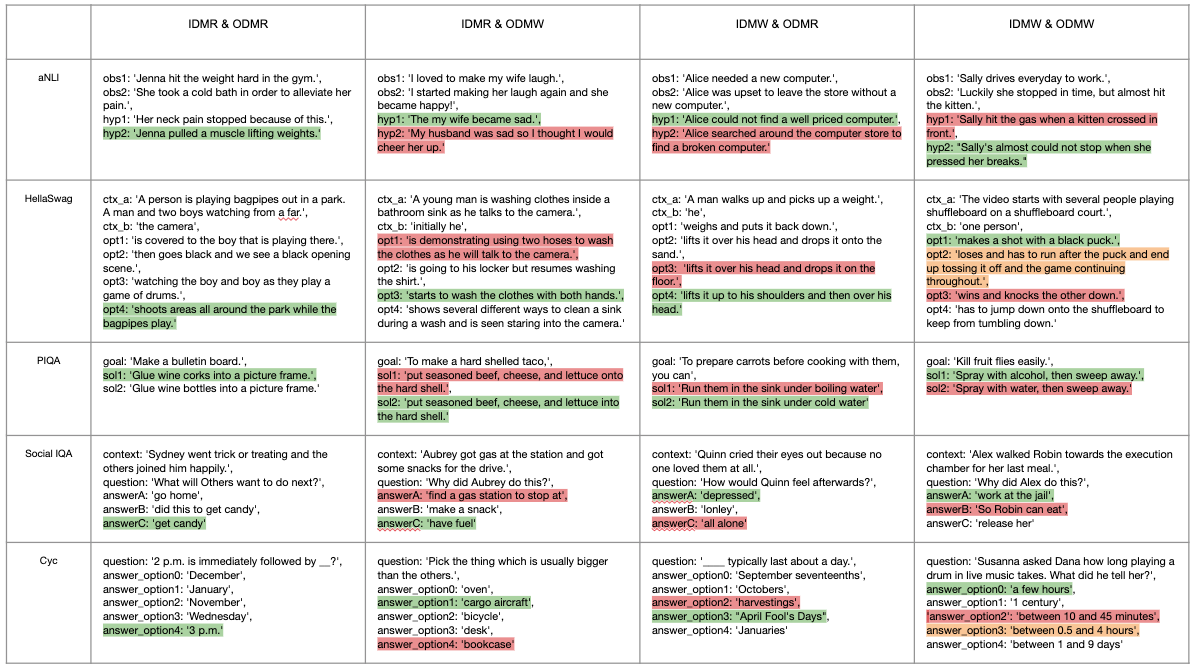}
\caption{Example instances for selected settings from Tables 5 and 6. IDMR and ODMR represent the population of instances where the in-domain model is right, and \emph{all} out-domain models are also right. Similarly, IDMW and ODMW represent the instances where the in-domain model was wrong, and all out-domain models retrieved the \emph{same} wrong answer. For each instance, the right and answers are highlighted in green and red, respectively. Note that, in the last column, it may be that the IDM wrong answer differs from the (same) wrong answer retrieved by the ODMs, in which case, we highlight the former in orange, and the latter in red.}
\label{fig:augseer}
\end{figure*}

{\bf Analysis of Incorrect/Correct Question Partitions.} While Tables 3 and 4 reveal that there are differences between in-domain and out-domain settings, it is not clear from that data where the different models are `going wrong'; specifically, where the in-domain training produces the right predictions, but an out-domain training setting does not. Table 5 produces the results of two \emph{consistency analyses}, with the top table  considering the benchmark-specific questions that the in-domain model (corresponding to that benchmark) got correct, and the bottom table for the questions that the in-domain model got wrong. Intriguingly, we find that, while relatively rare, there are questions (e.g., in the third row in the top table) that the in-domain model gets right but where \emph{all} out-domain models (ODMs) give the \emph{same} wrong answer. On the aNLI and PIQA \emph{dev.} sets, this situation is more common than the situation when all ODMs get the wrong answer, but at least two ODMs differ on the answer (the same situation arises in the bottom table). More research is required to investigate the deeper implications of this result, but one hypothesis is a `choice bias', at least in aNLI and PIQA. A choice bias arises when all answer choices do not have the same prior probability of being selected by the model. In particularly egregious cases, the question itself may not be necessary\footnote{This could arise, for example, if only one answer choice is grammatical, non-nonsensical, or both.} for the model to choose the right answer. A set of experimental studies, not dissimilar in design to the studies herein, could be used to prove or disprove the existence of a choice bias by not using the actual `question' (only the answer choices) during test-time (and subsequent to data collection, testing for significance against a random baseline). We leave such a study for future work. 

{\bf Qualitative Analysis.} To gain an `intuitive' sense of what the different out-domain models are learning, it is also instructive to consider samples of questions from all five benchmarks such that the in-domain model made the correct prediction but \emph{all} out-domain models made the \emph{same} wrong prediction. Figure 2 lists some examples of such questions. We find that, in some cases, the `wrong' answer is an `easy' wrong answer (yet either the in-domain or the out-domain model got it wrong). The example for aNLI in the third column (IDMW \& ODMR) is particularly telling. In other cases, the wrong answer is qualitatively `harder' (such as the example for Social IQA in the same column).  In yet other cases, we may be more inclined to agree with the machine learning models, such as for the Social IQA row and the last column. Some of the questions may require a degree of specialized or culture-specific knowledge (such as the HellaSwag question in the last column). 

The analysis here suggests several different aspects of an evaluation setup that are serving as `confounds' to testing not only the generalization of commonsense reasoning systems, but commonsense reasoning ability itself. Some of the questions in these benchmarks may not be commonsense questions, in the way that cognitive scientists have understood the phenomenon, and may have a domain- or culture-specific dependence that may cause noise in evaluations. However, in a few questions, we clearly see that the model gets some very obvious answers wrong. Therefore, it is not the case that the model only has trouble on difficult cases, though this also plays a role in the results that we observe. Addressing the flaws in these benchmarks, as well as using additional metrics (such as the performance loss) to evaluate candidate models, are both important avenues for future research. Eventually, a more advanced kind of benchmark may be warranted, one that has weighted or dynamically generated questions. Some researchers have suggested moving beyond multiple-choice questions altogether and focusing on generation tasks instead \cite{commongen}, \cite{DART}.

\section{Conclusion}
Language representation models such as BERT, RoBERTa and (more recently) GPT-3 have received prolific academic and media attention due to their ability to achieve near-human (or even, `super-human') performance on a range of individual benchmarks, including on several commonsense benchmarks. In this paper, we showed that there is still a significant performance drop when one such competitive model is trained on one kind of commonsense dataset but tested on another. It is important to remember that all datasets considered in  this work were supposed to test `commonsense reasoning', although some are more diverse and broader than others. The breadth of either the training or testing dataset is not found to significantly impact the overall conclusions. At minimum, our analyses suggest a potential source of dataset bias when evaluating commonsense reasoning. The large values of performance loss observed in  several settings strongly suggest that these commonsense models are not generalizing, and that more research is required before `human-like' commonsense reasoning performance can be confidently said to be within reach of these systems. 


\bibliographystyle{aaai}
\bibliography{references}

\end{document}